\documentclass[11pt]{article}

\usepackage{acl}

\usepackage{times}
\usepackage{latexsym}

\usepackage[T1]{fontenc}

\usepackage[utf8]{inputenc}

\usepackage{microtype}

\usepackage{inconsolata}

\usepackage{graphicx}
\usepackage{amssymb}

\usepackage{booktabs}
\usepackage{multirow}
\usepackage{booktabs}
\usepackage{multirow}
\usepackage[table]{xcolor}

\definecolor{CoolBlue}{HTML}{1E88E5}   
\definecolor{CoolGreen}{HTML}{00C853}  
\definecolor{CoolRed}{HTML}{FF1744}    

\usepackage{amsmath} 
\title{Radiology Report Generation for Low-Quality X-Ray Images}

\author{
Hongze Zhu\textsuperscript{1},
Yawen Huang\textsuperscript{2},
Jiaxuan Jiang\textsuperscript{1},
Chen Hu\textsuperscript{1},
Hong Liu\textsuperscript{2}, \\
\textbf{Ming Hu\textsuperscript{1},
Tianyu Wang\textsuperscript{1},
Zhijian Wu\textsuperscript{1,$\dagger$},
Yefeng Zheng\textsuperscript{1,$\dagger$}} \\
\textsuperscript{1}Westlake University \\
\textsuperscript{2}Tencent Jarvis Lab \\
\texttt{zhuhongze@westlake.edu.cn} \\
\texttt{wuzhijian@westlake.edu.cn, zhengyefeng@westlake.edu.cn} \\
\textsuperscript{$\dagger$} Corresponding authors
}

\begin{document}
\maketitle

\begin{abstract}
Vision-Language Models (VLMs) have significantly advanced automated Radiology Report Generation (RRG). However, existing methods implicitly assume high-quality inputs, overlooking the noise and artifacts prevalent in real-world clinical environments. Consequently, current models exhibit severe performance degradation when processing suboptimal images. To bridge this gap, we propose a robust report generation framework explicitly designed for image quality variations. We first introduce an Automated Quality Assessment Agent (AQAA) to identify low-quality samples within the MIMIC-CXR dataset and establish the Low-quality Radiology Report Generation (LRRG) benchmark. To tackle degradation-induced shifts, we propose a novel Dual-loop Training Strategy leveraging bi-level optimization and gradient consistency. This approach ensures the model learns quality-agnostic diagnostic features by aligning gradient directions across varying quality regimes. Extensive experiments demonstrate that our approach effectively mitigates model performance degradation caused by image quality deterioration. The code and data will be released upon acceptance.

\end{abstract}

\begin{figure}[t!]
\centering
\includegraphics[width=\columnwidth]{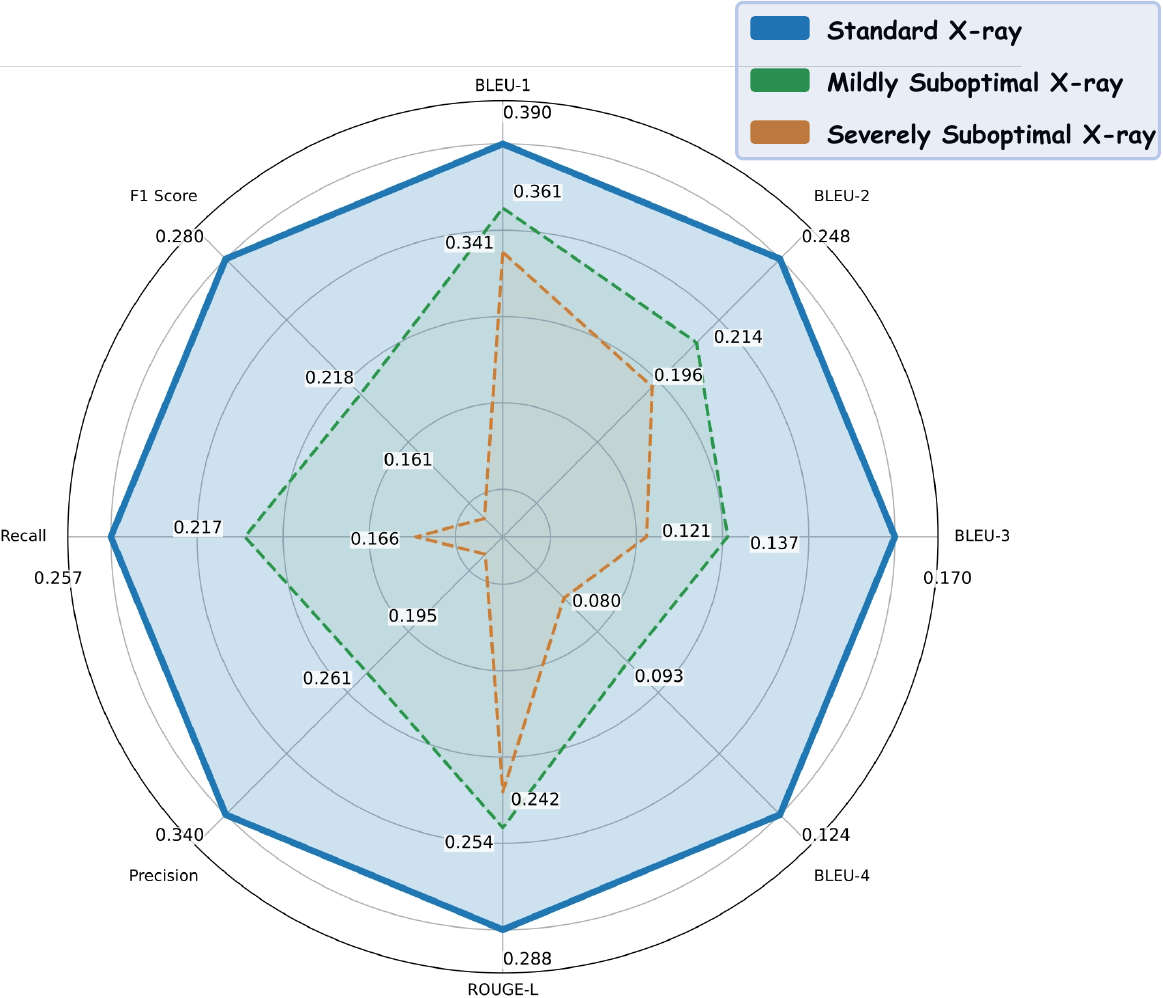}
\caption{Performance under quality degradation. We evaluate R2GenGPT~\cite{wang2023r2gengpt} across quality regimes and observe consistent declines in report generation performance as image quality degrades from \textit{standard} to \textit{mildly suboptimal} and \textit{severely suboptimal}.}
\label{fig0}
\end{figure}

\section{Introduction}

Chest X-ray (CXR) remains a key of clinical imaging because it is low-cost, readily available, and widely used in high-throughput settings such as emergency departments and intensive care units. 
Recent deep learning systems have achieved strong performance for CXR interpretation and have shown the potential to improve radiologists' accuracy in multireader evaluations \cite{rajpurkar2018deep,seah2021effect,cid2024development,dong2025keyword}. 
VLMs have significantly advanced automated RRG \cite{jin2025chain, wang2025cxpmrg}. Recent research focuses on workflow-aware clinician collaboration and the development of systematic modeling frameworks \cite{wang2023r2gengpt, tanno2025collaboration, wang2025survey, hou2025radar, zhang-etal-2025-libra}.

However, despite these achievements, existing literature overlooks a critical limitation: \textbf{the implicit assumption that input medical images are of consistently high quality.} In real-world clinical environments, image quality is frequently compromised by noise, motion blur, suboptimal exposure, patient positioning artifacts, and equipment-related distortions. These degradations significantly impair diagnostic utility and pose severe challenges for automated systems. Recent studies indicate that current deep learning models trained on high-quality datasets exhibit dramatic performance degradation when confronted with lower-quality images \cite{cheng2025understanding,chen2025noise,huang2025robust}. The problem of image quality variation is particularly acute in chest X-ray imaging, where technical factors such as variations in exposure parameters and grid misalignment, combined with patient-related factors like motion and body habitus, inevitably introduce variance.

Our preliminary investigations reveal that existing automated report generation systems, while performing admirably on standard benchmarks, exhibit substantial fragility when evaluated on lower-quality images, as shown in Fig. \ref{fig0}. This vulnerability raises serious concerns regarding the clinical applicability of these systems, as they may generate inaccurate or incomplete reports when processing suboptimal images, potentially leading to missed diagnoses or inappropriate clinical decisions.

To address this gap, we propose a novel approach to robust chest X-ray report generation that explicitly accounts for image quality variations. Our contributions are threefold. \textbf{First}, we developed an AQAA based on the Qwen-VL-30B \cite{yang2025qwen3} to identify and stratify low-quality image-report pairs from the MIMIC-CXR dataset \cite{johnson2019mimic}. Our quality assessment framework combines multiple image quality metrics to create a comprehensive quality scoring system. \textbf{Second}, based on the proposed AQAA, we introduce the LRRG task, a new benchmark specifically designed to evaluate the robustness of automated report generation systems under realistic image quality conditions. This benchmark addresses the limitations of existing datasets that predominantly feature high-quality images, thereby providing a more realistic assessment of system performance in clinical environments. \textbf{Third}, we propose a novel Dual-loop Training Strategy that leverages bi-level optimization to achieve quality-agnostic report generation. By implementing a regime rotation scheme with gradient consistency regularization, our method ensures that the descent directions on varying quality regimes are coherent. This effectively prevents the model from relying on unstable, quality-dependent features, thereby guaranteeing reliable diagnostic outputs for degradation scenarios. In summary, our contributions are as follows:

\vspace{-0.7em} \begin{itemize} \setlength{\itemsep}{0pt} \setlength{\parsep}{0pt} \setlength{\parskip}{0pt} 
\item We developed an AQAA based on MLLM to systematically identify and stratify low-quality images in the MIMIC-CXR dataset. 
\item We introduced the LRRG task, establishing a rigorous benchmark to evaluate model robustness against real-world image degradation. 
\item We proposed a bi-level optimization strategy using regime rotation and gradient consistency. This ensures the model learns diagnostic features that remain invariant across quality shifts. 
\item Extensive experiments demonstrate that our approach significantly outperforms state-of-the-art models, delivering reliable reports for degraded inputs. 
\end{itemize}

\begin{figure*}[t]
\centerline{\includegraphics[width=1.02\textwidth]{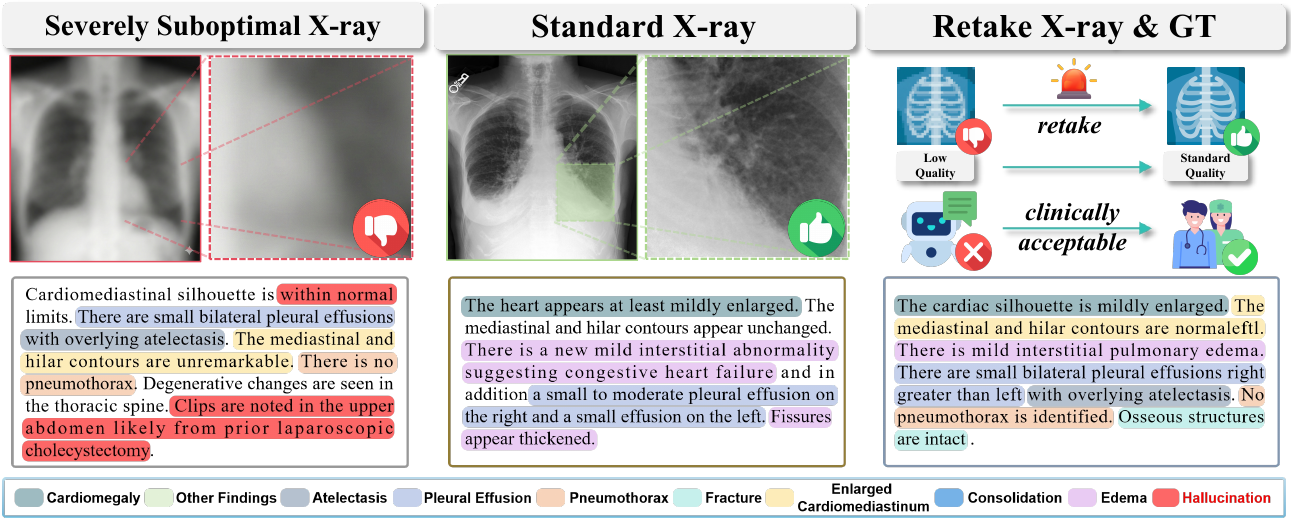}}
\caption{Retake pairs illustrate acquisition-related quality shift. A low-quality pre-retake CXR can degrade report generation, while the post-retake CXR improves alignment with the reference report.}
\label{fig1}
\end{figure*}

\section{Related Work}
Early RRG systems incorporated clinical semantics into standard encoder-decoder architectures~\cite{chen2020generating, liu2021exploring, wang2023metransformer, tanida2023interactive, gu2025orid, jeong2024multimodal}. 
The advent of foundation models has catalyzed a paradigm shift from specialized encoder-decoder pipelines to generalist Multimodal Large Language Model (MLLM) frameworks. 
For instance, R2GenGPT~\cite{wang2023r2gengpt} bridges visual encoders with frozen LLMs via efficient alignment to minimize training costs. 
Concurrently, domain-specific MLLMs like LLaVA-Med~\cite{li2023llava} and BiomedGPT~\cite{luo2024biomedgpt} have unlocked impressive zero-shot capabilities and instruction-following skills for report drafting. 
Most recently, this scope has expanded further with M$^3$FM~\cite{liu2025multimodal}, a unified foundation model that supports zero-shot diagnosis across diverse imaging modalities and languages. 
Beyond static generation, emerging interactive systems such as Flamingo-CXR~\cite{tanno2025collaboration} reveal that automatic scores often diverge from clinical reality, necessitating evaluation protocols grounded in radiologist workflows.

Current models have mastered complex reasoning. 
However, they overlook a critical clinical reality. 
Image quality often drops as patient severity rises. 
The most critical ICU cases inevitably yield the noisiest scans. 
If we ignore this, models risk failing precisely when help is needed most. 
Addressing this is not just technical progress. 
It is the foundation for safe and responsible clinical AI.

\section{Data Stratification}
To systematically study the impact of image degradation, we require a dataset that reflects real-world clinical shifts rather than synthetic noise. We address the lack of explicit quality annotations in standard benchmarks by establishing a multi-stage stratification framework.



\subsection{The Gold Standard: The Retake Pairs}
\label{sec:retake}

In routine clinical workflows, technicians review images immediately after acquisition.
If quality is insufficient due to motion, underexposure, or positioning errors, a retake is triggered.
This process creates a natural paired sample: a pre-retake low-quality image and a post-retake high-quality reference. We leverage this phenomenon to establish a rigorous Gold Standard for validating our quality assessment agent.
Extracting these events from MIMIC-CXR is non-trivial due to redundant duplicates and functional comparisons. We implemented a strict curation pipeline:

\noindent
\textbf{1) Temporal \& Spatial Alignment:} We isolated consecutive scans within a 30-minute window, utilizing DICOM tags to ensure consistent projection (AP/PA) and excluding position changes.

\noindent
\textbf{2) Content Consistency:} To prevent image stitching artifacts, we discarded pairs with low field-of-view overlap (IoU $< 0.7$).

\noindent
\textbf{3) Semantic Filtering:} An LLM screened metadata to remove functional exams, followed by manual validation by board-certified radiologists.

The resulting 100 high-confidence pairs constitute our \textit{Retake Standard}, serving as the ground truth for aligning our agent with human judgment.

\begin{figure*}[t!]
\centerline{\includegraphics[width=1.02\textwidth]{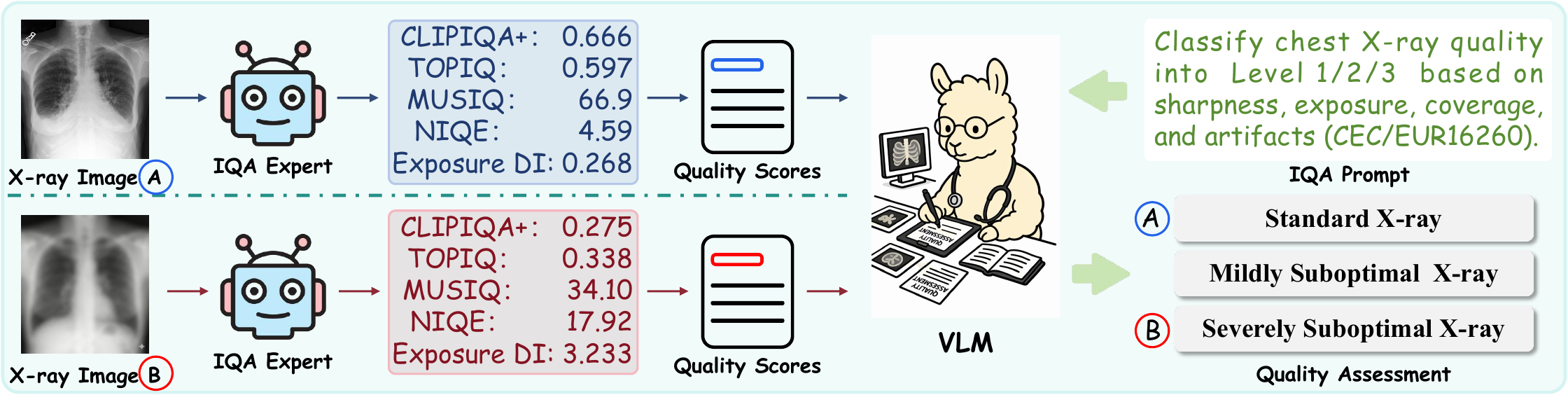}}
\caption{Overview of our quality assessment agent. We first extract no-reference IQA scores and an exposure deviation index from each CXR, then feed these priors with a CEC/EUR16260-aligned prompt into a VLM to classify image quality into three levels: standard, mildly suboptimal, and severely suboptimal.}
\label{fig3}
\end{figure*}

\subsection{Automated Quality Assessment Agent}

To scale quality annotation beyond the limited retake pairs, we develop the AQAA. This module bridges the gap between low-level pixel statistics and high-level clinical interpretation by integrating expert priors into a Multimodal large language model Qwen-VL-30B \cite{yang2025qwen3}.

\paragraph{Physics-based Exposure Anchor.}
To ground the assessment in physical reality, we extract the \textbf{Deviation Index (DI)} \cite{fauber2020radiographic} directly from the DICOM metadata (Tag 0018,1411).
Unlike visual features, the DI provides an objective measure of exposure accuracy relative to the detector calibration.
It quantifies the deviation from the target exposure index ($EI_T$).
This serves as a physics-based anchor that is independent of image post-processing or display adjustments.

\paragraph{Perceptual Quality Priors.}
We complement the physical DI with four heterogeneous Image Quality Assessment (IQA) metrics.
These metrics capture degradation patterns across different granularities.
\textbf{NIQE} \cite{zhang2015feature} utilizes natural scene statistics to quantify distributional shifts and noise without reference images.
\textbf{MUSIQ} \cite{ke2021musiq} employs a multi-scale transformer architecture for handling the varying resolutions radiography.
\textbf{CLIPIQA} \cite{wang2023exploring} and \textbf{TOPIQ} \cite{chen2024topiq} leverage Contrastive Language-Image Pre-training (CLIP) \cite{radford2021learning}.
These two metrics evaluate the preservation of semantic content and textural details through text-image alignment scores.

\paragraph{Criteria-Guided Inference.}
The inference mechanism is formulated as a standard-aligned reasoning task.
We engineered the prompt to adhere to the \textit{CEC EUR 16260} protocol.
It directs the VLM to evaluate anatomical landmarks and exposure adequacy under this standardized framework.
Simultaneously, the encoded metrics $(DI, Q_{NIQE}, Q_{MUSIQ}, Q_{CLIP}, Q_{TOP})$ serve as objective constraints within the textual context.
By conditioning on both clinical guidelines and statistical priors, the VLM reconciles semantic observations with technical rigor.
This synthesis yields a classification ranging across Levels~1--3.

\begin{table}[t]
\centering
\caption{Distribution of labels in Mildly ($N=32,826$) and Severely ($N=22,840$) Suboptimal datasets.}
\label{tab:data_distribution}
\resizebox{\columnwidth}{!}{%
\begin{tabular}{l cc}
\toprule
\textbf{Condition} &
\textbf{Mildly ($N=32{,}826$)} &
\textbf{Severely ($N=22{,}840$)} \\
\midrule
Enlarged  Cardiomediastinum & 2,019 (6.15\%) & 1,481 (6.48\%) \\
Cardiomegaly      & 9,631 (29.34\%) & 6,471 (28.33\%) \\
Lung Opacity      & 11,055 (33.68\%) & 8,870 (38.84\%) \\
Lung Lesion       & 1,044 (3.18\%) & 647 (2.83\%) \\
Edema             & 5,126 (15.62\%) & 3,974 (17.40\%) \\
Consolidation     & 1,569 (4.78\%) & 1,343 (5.88\%) \\
Pneumonia         & 1,169 (3.56\%) & 900 (3.94\%) \\
Atelectasis       & 10,644 (32.43\%) & 7,603 (33.29\%) \\
Pneumothorax      & 1,662 (5.06\%) & 1,214 (5.32\%) \\
Pleural Effusion  & 9,541 (29.07\%) & 7,918 (34.67\%) \\
Pleural Other     & 388 (1.18\%) & 304 (1.33\%) \\
Fracture          & 1,049 (3.20\%) & 790 (3.46\%) \\
Support Devices   & 16,868 (51.39\%) & 14,399 (63.04\%) \\
No Finding        & 4,582 (13.96\%) & 2,536 (11.10\%) \\
\bottomrule
\end{tabular}
}
\end{table}

\subsection{Constructing Graded Quality Regimes}

Leveraging the AQAA, we restructured the MIMIC-CXR dataset into three distinct regimes to simulate varying degrees of clinical degradation.
First, we isolated standard frontal-view examinations (PA/AP) acquired with stationary machines to form the \textbf{Standard Regime}.
This subset represents the ideal distribution of high-quality clinical data ($\mathcal{D}_{std}$).
Second, for low-quality X-ray scans, we use the agent's Level 2 and Level 3 predictions to form two graded sets: \textbf{mildly suboptimal} ($\mathcal{D}_{mild}$) and \textbf{severely suboptimal} ($\mathcal{D}_{severe}$).

\paragraph{Strict Patient-Level Isolation.}
A critical challenge in retrospective analysis is patient overlap, where scans of the same patient appear in both training and testing sets, leading to inflated performance estimates. To ensure rigorous evaluation, we implemented a \textit{leakage-proof splitting strategy}:
    \textbf{1) Test Set Lockdown:} We first randomly sampled 500 distinct exams from both the $\mathcal{D}_{mild}$ and $\mathcal{D}_{severe}$ pools to form the fixed evaluation benchmarks.
  \textbf{2) Decontamination:} Any subject (patient ID) appearing in these test sets was rigorously purged from all training partitions. 

This process guarantees that the model is evaluated on strictly unseen patients, providing an unbiased assessment of generalization.

\begin{table}[t]
\centering
\small
\caption{Statistics of the constructed quality regimes. $\mathcal{D}_{std}$ represents ideal stationary scans. $\mathcal{D}_{mild}$ and $\mathcal{D}_{severe}$ denote scans with graded quality degradation.}
\label{tab:dataset_stats}
\resizebox{\linewidth}{!}{%
\begin{tabular}{l rrr}
\toprule
\multirow{2}{*}{\textbf{Quality Regime}} & \multicolumn{3}{c}{\textbf{Partition Statistics}} \\
\cmidrule{2-4}
 & \textbf{Train} & \textbf{Val} & \textbf{Test} \\ 
\midrule
Standard ($\mathcal{D}_{std}$) & 104,827 & 857 & 1,780 \\
Mildly Suboptimal ($\mathcal{D}_{mild}$) & 28,791 & 240 & 515 \\
Severely Suboptimal ($\mathcal{D}_{severe}$) & 19,189 & 186 & 514 \\ 
\textit{Auxiliary Low-Quality} & -- & -- & 669 \\ 
\midrule
\textbf{Total} & \textbf{152,807} & \textbf{1,283} & \textbf{3,478} \\
\bottomrule
\end{tabular}%
}
\end{table}

\paragraph{Validation against Gold Standard.}
Before large-scale application, we validated AQAA against our Retake Standard (Sec. \ref{sec:retake}). We defined a consistency metric where the agent passes if it rates the pre-retake (low-quality) image equal to or lower than the post-retake reference. The agent achieved a \textbf{99\% consistency rate}, demonstrating alignment with physical quality improvements.

Applying this verified agent to the full MIMIC-CXR database yielded our final graded regimes, as shown in Table~\ref{tab:data_distribution}. Quantitative analysis reveals that the Severely Suboptimal regime exhibits a higher prevalence of critical findings compared to the Mildly Suboptimal group. Specifically, the proportion of Support Devices rises from 51.39\% to 63.04\% and Lung Opacity increases from 33.68\% to 38.84\%, concurrent with a decrease in No Finding cases from 13.96\% to 11.10\%. This distribution shift mirrors the clinical reality where critically ill patients often require bedside imaging under difficult acquisition conditions. Consequently, the increased presence of severity indicators further validates that our AQAA successfully captures authentic, severity-correlated quality degradation.

As detailed in Table~\ref{tab:dataset_stats}, $\mathcal{D}_{std}$ serves as the high-quality anchor with over 100k training samples.
$\mathcal{D}_{mild}$ and $\mathcal{D}_{severe}$ provide substantial volumes of 28,791 and 19,189 training samples, respectively.
Regarding evaluation, in addition to our strictly isolated test sets ($N \approx 500$ each), we retained 669 naturally occurring low-quality images from the official MIMIC-CXR test set to serve as an auxiliary benchmark.

\begin{figure*}[t]
\centerline{\includegraphics[width=1.02\textwidth]{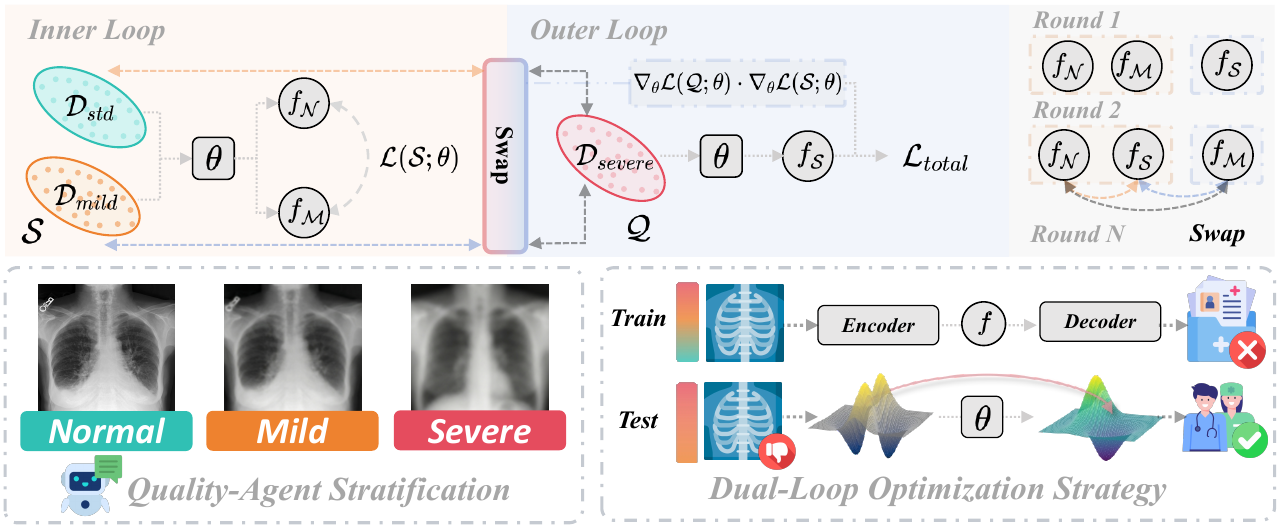}}
\caption{Overview of our quality assessment agent. We compute no-reference IQA signals and an exposure deviation index for each CXR, then use a VLM with a CEC/EUR16260-aligned prompt to grade images into three levels: standard, mildly suboptimal, and severely suboptimal.}
\label{fig4}
\end{figure*}

\section{Dual-loop Training Strategy}

Existing RRG models typically assume stable image quality, leading to severe performance degradation when deployed on low-quality X-ray acquisitions. This vulnerability arises because naive mixing of data from different quality regimes often encourages models to exploit quality-specific spurious cues rather than diagnostic evidence \cite{Li2018MLDG}. To address this issue, we propose a \textbf{Dual-loop Training Strategy} that aligns representations across quality regimes via inner–outer optimization, regime rotation, and gradient coherence regularization, improving diagnostic consistency under severe degradations.

\subsection{Methodology}
\label{sec:methodology}

We formulate robustness to image quality variation as a bi-level optimization problem rather than a domain adaptation task. Specifically, our objective is to learn a parameter state $\theta$ such that the descent direction induced by high-quality data remains consistent with that induced by low-quality data.

\paragraph{Regime Partitioning.}
We consider three quality regimes: $\mathcal{D}_{std}$, $\mathcal{D}_{mild}$, and $\mathcal{D}_{severe}$.
At each iteration, we randomly partition these regimes into a \textit{support set} $\mathcal{S}$ for the inner loop optimization and a \textit{query set} $\mathcal{Q}$ for the outer loop evaluation.
This randomized regime rotation prevents overfitting to any single quality condition and encourages the learning of quality-invariant representations.

\paragraph{Inner Loop: Virtual Adaptation.}
We first simulate a training step on $\mathcal{S}$.
This process computes a temporary parameter state $\tilde{\theta}$ that is virtually adapted to the current visible quality distributions.
Formally, the inner update is calculated as
\begin{equation}
    \tilde{\theta} = \theta - \alpha \nabla_{\theta} \mathcal{L}(\mathcal{S}; \theta)
\end{equation}
where $\alpha$ denotes the inner step size and $\mathcal{L}$ represents the report generation loss.
This step essentially projects a virtual trajectory towards minimizing the loss of the support regimes.

\paragraph{Outer Loop: Gradient Coherence Optimization.}
To improve robustness across quality regimes, we require that updates computed from $\mathcal{S}$ also generalize to the unseen $\mathcal{Q}$.
Accordingly, the outer-loop objective minimizes the query loss evaluated at the adapted parameters $\tilde{\theta}$:
\begin{equation}
    \min_{\theta} \quad \mathcal{L}(\mathcal{Q}; \tilde{\theta}) = \mathcal{L}(\mathcal{Q}; \theta - \alpha \nabla_{\theta} \mathcal{L}(\mathcal{S}; \theta))
\end{equation}
To make the interaction between quality regimes explicit, we apply a first-order Taylor expansion to the above objective, yielding the approximation:
{\small
\begin{equation}
    \mathcal{L}_{total} \approx \mathcal{L}(\mathcal{Q}; \theta) - \alpha \underbrace{ (\nabla_{\theta} \mathcal{L}(\mathcal{Q}; \theta) \cdot \nabla_{\theta} \mathcal{L}(\mathcal{S}; \theta)) }_{\text{Gradient Coherence}}
\end{equation}
}The second term in Eq. (3) encourages alignment between the descent directions induced by the support and query regimes.
It therefore acts as a \textit{coherence regularizer}, discouraging conflicting or orthogonal updates that arise from quality-specific artifacts.
As a result, the optimization is biased toward representations that remain consistent across varying acquisition qualities.

\subsection{Theoretical Analysis}
\label{sec:analysis}

In this section, we derive the implicit regularization effect of the proposed dual-loop optimization. The key idea is to treat the inner-loop update as a small perturbation of the parameters and analyze how it changes the outer objective via a Taylor expansion.

\paragraph{Bi-level Objective.}
Let $\mathcal{D}$ denote a distribution over quality regimes. At each iteration, we sample two disjoint sets, a support set $\mathcal{S}$ and a query $\mathcal{Q}$ set. The dual-loop objective minimizes the expected query loss evaluated at parameters adapted on $\mathcal{S}$:
\begin{equation}
    \min_{\theta} \mathcal{J}(\theta) = \mathbb{E}_{\mathcal{S}, \mathcal{Q} \sim \mathcal{D}} \left[ \mathcal{L}(\mathcal{Q}; \phi(\theta)) \right]
\end{equation}
where the inner adaptation map is 
\begin{equation}
\phi(\theta) = \theta - \alpha \nabla_{\theta} \mathcal{L}(\mathcal{S}; \theta),
\end{equation} 
and $\alpha$ is the step size.

\paragraph{Taylor Expansion and Curvature Analysis.}
To expose the implicit regularization effect induced by the inner update, we perform a local Taylor expansion of the query loss around the current parameters $\theta$.
Let $\mathbf{g}_{\mathcal{S}} = \nabla_{\theta} \mathcal{L}(\mathcal{S}; \theta)$ and $\mathbf{g}_{\mathcal{Q}} = \nabla_{\theta} \mathcal{L}(\mathcal{Q}; \theta)$ denote the gradients on the support and query sets, respectively.
The inner update is a displacement $\Delta\theta = -\alpha \mathbf{g}_{\mathcal{S}}$.
For sufficiently small $\alpha$, we have
\begin{equation}
\begin{aligned}
\mathcal{J}(\theta)
&= \mathcal{L}(\mathcal{Q}; \theta + \Delta\theta) \\
&\approx \mathcal{L}(\mathcal{Q}; \theta)
+ \mathbf{g}_{\mathcal{Q}}^\top \Delta\theta
+ \mathcal{O}(\|\Delta\theta\|^2) \\
&= \mathcal{L}(\mathcal{Q}; \theta)
- \alpha \langle \mathbf{g}_{\mathcal{Q}}, \mathbf{g}_{\mathcal{S}} \rangle
+ \mathcal{O}(\alpha^2).
\end{aligned}
\end{equation}

\paragraph{Derivation of Gradient Coherence.}
The leading interaction term is the inner product between gradients from the two regimes:
{\small
\begin{equation}
\mathcal{R}_{\text{cohere}}
= - \alpha \langle \mathbf{g}_{\mathcal{Q}}, \mathbf{g}_{\mathcal{S}} \rangle
= - \alpha \|\mathbf{g}_{\mathcal{Q}}\|\, \|\mathbf{g}_{\mathcal{S}}\| \cos(\varphi),
\end{equation}
}where $\varphi$ is the angle between the two descent directions.
Minimizing $\mathcal{J}(\theta)$ thus encourages $\cos(\varphi)$ to be large and positive, \textit{i.e.}, aligned updates across quality regimes.

\paragraph{Geometric Interpretation: Orthogonal Noise Filtering.}
The above analysis highlights a key difference from standard Empirical Risk Minimization (ERM).
ERM optimizes the expected loss over pooled data from mixed quality regimes, without explicitly constraining how gradients from different regimes interact.
In contrast, our objective includes the coherence term
$\mathcal{R}_{\text{cohere}} = -\alpha \langle \mathbf{g}_{\mathcal{Q}}, \mathbf{g}_{\mathcal{S}} \rangle$,
which imposes a directional constraint on the updates:
\begin{itemize}
    \item \textbf{Consistency reward:} if $ \cos(\phi) > 0 $ (aligned gradients), then
    $ \langle \mathbf{g}_{\mathcal{Q}}, \mathbf{g}_{\mathcal{S}} \rangle > 0 $,
    and the coherence term decreases the objective, reinforcing shared descent directions.
    \item \textbf{Conflict penalty:} if $ \cos(\phi) < 0 $ (conflicting gradients), then
    $ \langle \mathbf{g}_{\mathcal{Q}}, \mathbf{g}_{\mathcal{S}} \rangle < 0 $,
    and the coherence term increases the objective, discouraging regime-specific update directions.
\end{itemize}
In radiology images, pathology-related gradients are consistent across regimes, while noise-related gradients tend to be orthogonal or conflicting across varying acquisition qualities. Maximizing gradient coherence, therefore, biases optimization toward shared diagnostic directions and suppresses regime-specific spurious components.

\begin{table*}[!t] 
\centering
\caption{Performance comparison on the four benchmarks. CE Metrics now encompass advanced clinical alignment scores. Where \textbf{sft} denotes supervised fine-tuning, and \textbf{DTS} refers to the proposed dual-loop training strategy. We adopt the color scheme: \colorbox[HTML]{BDE6CD}{\textbf{Best}} and \colorbox[HTML]{E4EEBC}{Second Best}.}
\label{tab:main_results}
\resizebox{\textwidth}{!}{%
\begin{tabular}{@{}ll cccc ccccc@{}}
\toprule
 &  & \multicolumn{4}{c}{NLG METRICS} & \multicolumn{5}{c}{CE METRICS} \\ 
\cmidrule(lr){3-6} \cmidrule(l){7-11}
\multirow{-2}{*}{Benchmarks} & \multirow{-2}{*}{Methods} & BLEU-1 & BLEU-4 & METEOR & ROUGE-L & Precision & Recall & F1 Score & RaTEScore & RadGraph \\ 
\midrule

\multirow{9}{*}{$T_{std}$} & GPT5.2 \cite{achiam2023gpt} 
& 0.137 & 0.006 & 0.108 & 0.109 & 0.215 & 0.203 & 0.185 & 0.481 & 0.180 \\
& EKAGen \cite{bu2024instance} 
& \cellcolor[HTML]{BDE6CD}\textbf{0.422} & 0.120 & 0.158 & 0.271 
& \cellcolor[HTML]{BDE6CD}\textbf{0.385} & \cellcolor[HTML]{BDE6CD}\textbf{0.278} & \cellcolor[HTML]{BDE6CD}\textbf{0.294} & 0.534 
& \cellcolor[HTML]{BDE6CD}\textbf{0.284} \\
& Mamba\_Xray \cite{wang2025cxpmrg} 
& 0.398 & 0.121 & \cellcolor[HTML]{E4EEBC}0.172 & 0.270 
& 0.331 & 0.258 & \cellcolor[HTML]{E4EEBC}0.277 & \cellcolor[HTML]{E4EEBC}0.540 & \cellcolor[HTML]{E4EEBC}0.280 \\
& Qwen3VL4B \cite{yang2025qwen3} 
& 0.087 & 0.009 & 0.111 & 0.115 & 0.215 & 0.228 & 0.183 & 0.518 & 0.155 \\
& Qwen3VL4B-sft \cite{yang2025qwen3} 
& 0.243 & 0.063 & 0.141 & 0.212 & 0.332 & 0.247 & 0.252 & 0.538 & 0.271 \\
& Qwen3VL8B \cite{yang2025qwen3} 
& 0.094 & 0.008 & 0.111 & 0.114 & 0.226 & 0.163 & 0.163 & 0.518 & 0.167 \\
& Qwen3VL8B-sft \cite{yang2025qwen3} 
& 0.236 & 0.061 & 0.139 & 0.212 & \cellcolor[HTML]{E4EEBC}0.358 & 0.233 & 0.243 & 0.536 & 0.266 \\
& R2GenGPT~\cite{wang2023r2gengpt} 
& 0.390 & \cellcolor[HTML]{E4EEBC}0.123 & \cellcolor[HTML]{BDE6CD}\textbf{0.174} & \cellcolor[HTML]{E4EEBC}0.288 
& 0.337 & 0.254 & 0.275 & \cellcolor[HTML]{BDE6CD}\textbf{0.542} & 0.279 \\
& R2GenGPT-DTS (Ours) 
& \cellcolor[HTML]{E4EEBC}0.418 & \cellcolor[HTML]{BDE6CD}\textbf{0.131} & 0.167 & \cellcolor[HTML]{BDE6CD}\textbf{0.291} 
& 0.334 & \cellcolor[HTML]{E4EEBC}0.259 & \cellcolor[HTML]{E4EEBC}0.277 & \cellcolor[HTML]{BDE6CD}\textbf{0.542} & \cellcolor[HTML]{E4EEBC}0.280 \\
\midrule

\multirow{7}{*}{$T_{mild}$} & GPT5.2 \cite{achiam2023gpt} 
& 0.128 & 0.007 & 0.104 & 0.108 & 0.225 & 0.238 & 0.197 & 0.473 & 0.167 \\
& Qwen3VL4B \cite{yang2025qwen3} 
& 0.071 & 0.007 & 0.102 & 0.099 & 0.211 & 0.273 & 0.205 & 0.499 & 0.137 \\
& Qwen3VL4B-sft \cite{yang2025qwen3} 
& 0.222 & 0.051 & 0.132 & 0.199 & 0.349 & 0.256 & 0.269 & 0.521 & \cellcolor[HTML]{E4EEBC}0.257 \\
& Qwen3VL8B \cite{yang2025qwen3} 
& 0.074 & 0.005 & 0.102 & 0.095 & 0.181 & 0.169 & 0.153 & 0.497 & 0.143 \\
& Qwen3VL8B-sft \cite{yang2025qwen3} 
& 0.219 & 0.048 & 0.128 & 0.193 & \cellcolor[HTML]{E4EEBC}0.366 & 0.224 & 0.239 & 0.509 & 0.243 \\
& R2GenGPT~\cite{wang2023r2gengpt} 
& \cellcolor[HTML]{E4EEBC}0.338 & \cellcolor[HTML]{E4EEBC}0.092 & \cellcolor[HTML]{BDE6CD}\textbf{0.166} & \cellcolor[HTML]{E4EEBC}0.260 
& 0.353 & \cellcolor[HTML]{E4EEBC}0.288 & \cellcolor[HTML]{E4EEBC}0.291 & \cellcolor[HTML]{E4EEBC}0.524 & 0.255 \\
& R2GenGPT-DTS (Ours) 
& \cellcolor[HTML]{BDE6CD}\textbf{0.376} & \cellcolor[HTML]{BDE6CD}\textbf{0.103} & \cellcolor[HTML]{E4EEBC}0.161 & \cellcolor[HTML]{BDE6CD}\textbf{0.271} 
& \cellcolor[HTML]{BDE6CD}\textbf{0.411} & \cellcolor[HTML]{BDE6CD}\textbf{0.298} & \cellcolor[HTML]{BDE6CD}\textbf{0.321} & \cellcolor[HTML]{BDE6CD}\textbf{0.530} & \cellcolor[HTML]{BDE6CD}\textbf{0.269} \\
\midrule

\multirow{7}{*}{$T_{severe}$} & GPT5.2 \cite{achiam2023gpt} 
& 0.130 & 0.007 & 0.104 & 0.105 & 0.210 & 0.209 & 0.177 & 0.472 & 0.162 \\
& Qwen3VL4B \cite{yang2025qwen3} 
& 0.076 & 0.007 & 0.105 & 0.102 & 0.225 & \cellcolor[HTML]{E4EEBC}0.273 & 0.208 & 0.506 & 0.144 \\
& Qwen3VL4B-sft \cite{yang2025qwen3} 
& 0.246 & 0.052 & 0.130 & 0.199 & \cellcolor[HTML]{E4EEBC}0.354 & 0.263 & 0.274 & 0.527 & 0.250 \\
& Qwen3VL8B \cite{yang2025qwen3} 
& 0.079 & 0.006 & 0.106 & 0.098 & 0.218 & 0.203 & 0.185 & 0.498 & 0.146 \\
& Qwen3VL8B-sft \cite{yang2025qwen3} 
& 0.236 & 0.052 & 0.130 & 0.198 & \cellcolor[HTML]{BDE6CD}\textbf{0.373} & 0.265 & \cellcolor[HTML]{E4EEBC}0.275 & 0.521 & 0.248 \\
& R2GenGPT~\cite{wang2023r2gengpt} 
& \cellcolor[HTML]{E4EEBC}0.367 & \cellcolor[HTML]{E4EEBC}0.095 & \cellcolor[HTML]{E4EEBC}0.159 & \cellcolor[HTML]{E4EEBC}0.265 
& 0.320 & \cellcolor[HTML]{BDE6CD}\textbf{0.274} & \cellcolor[HTML]{E4EEBC}0.275 & \cellcolor[HTML]{E4EEBC}0.535 & \cellcolor[HTML]{E4EEBC}0.269 \\
& R2GenGPT-DTS (Ours) 
& \cellcolor[HTML]{BDE6CD}\textbf{0.383} & \cellcolor[HTML]{BDE6CD}\textbf{0.103} & \cellcolor[HTML]{BDE6CD}\textbf{0.160} & \cellcolor[HTML]{BDE6CD}\textbf{0.268} 
& 0.323 & \cellcolor[HTML]{BDE6CD}\textbf{0.274} & \cellcolor[HTML]{BDE6CD}\textbf{0.277} & \cellcolor[HTML]{BDE6CD}\textbf{0.536} & \cellcolor[HTML]{BDE6CD}\textbf{0.275} \\
\midrule

\multirow{9}{*}{$T_{aux}$} & GPT5.2 \cite{achiam2023gpt} 
& 0.137 & 0.005 & 0.099 & 0.107 & 0.218 & 0.218 & 0.186 & 0.468 & 0.153 \\
& EKAGen \cite{bu2024instance} 
& \cellcolor[HTML]{BDE6CD}\textbf{0.390} & \cellcolor[HTML]{E4EEBC}0.103 & 0.154 & 0.259 
& \cellcolor[HTML]{BDE6CD}\textbf{0.437} & 0.273 & \cellcolor[HTML]{E4EEBC}0.299 & 0.503 & \cellcolor[HTML]{E4EEBC}0.266 \\
& Mamba\_Xray \cite{wang2025cxpmrg} 
& 0.361 & 0.091 & \cellcolor[HTML]{BDE6CD}\textbf{0.160} & 0.247 
& 0.387 & 0.272 & 0.278 & 0.514 & 0.254 \\
& Qwen3VL4B \cite{yang2025qwen3}  
& 0.077 & 0.007 & 0.104 & 0.104 & 0.223 & 0.275 & 0.208 & 0.498 & 0.137 \\
& Qwen3VL4B-sft \cite{yang2025qwen3} 
& 0.237 & 0.054 & 0.131 & 0.198 & 0.338 & \cellcolor[HTML]{E4EEBC}0.284 & 0.286 & 0.516 & 0.255 \\
& Qwen3VL8B \cite{yang2025qwen3} 
& 0.081 & 0.006 & 0.106 & 0.101 & 0.198 & 0.201 & 0.182 & 0.493 & 0.140 \\
& Qwen3VL8B-sft \cite{yang2025qwen3} 
& 0.225 & 0.054 & 0.131 & 0.199 & 0.335 & 0.238 & 0.255 & 0.505 & 0.243 \\
& R2GenGPT~\cite{wang2023r2gengpt} 
& 0.383 & \cellcolor[HTML]{BDE6CD}\textbf{0.107} & 0.157 & \cellcolor[HTML]{BDE6CD}\textbf{0.271} 
& 0.329 & 0.280 & 0.286 & \cellcolor[HTML]{E4EEBC}0.519 & 0.263 \\
& R2GenGPT-DTS (Ours) 
& \cellcolor[HTML]{E4EEBC}0.388 & \cellcolor[HTML]{E4EEBC}0.104 & \cellcolor[HTML]{E4EEBC}0.159 & \cellcolor[HTML]{E4EEBC}0.269 
& \cellcolor[HTML]{E4EEBC}0.388 & \cellcolor[HTML]{BDE6CD}\textbf{0.293} & \cellcolor[HTML]{BDE6CD}\textbf{0.307} & \cellcolor[HTML]{BDE6CD}\textbf{0.523} & \cellcolor[HTML]{BDE6CD}\textbf{0.267} \\
\bottomrule
\end{tabular}%
}
\end{table*}

\section{Experiments}

\subsection{Task and Dataset}
\label{subsec:task_and_dataset}
\paragraph{Task Definition.}
We focus on the task of LRRG from single-view radiographs.
Formally, given a current frontal-view image $I$, the model generates a report $R$.
To isolate image quality as the sole variable, we impose two strict constraints:
(1) Input is restricted to single-view frontal scans (AP/PA) to eliminate confounding factors from lateral views;
(2) \textbf{No historical information} is provided.
The model must rely exclusively on the visual evidence present in the current acquisition, preventing information leakage from prior exams.

\paragraph{Evaluation Benchmarks.}
We established four testing benchmarks ($T$), each representing a distinct clinical acquisition scenario:

\noindent\textbf{1) Standard Benchmark ($T_{std}$):}
Comprising 1,780 high-quality images from the Standard Regime.
This set establishes the performance baseline under ideal, stationary acquisition conditions.

\noindent\textbf{2) Mildly Suboptimal Benchmark ($T_{mild}$):}
Consisting of 515 Level 2 images.
This set represents \textit{routine technical deviations}, such as slight positioning errors or minor motion.
It evaluates model stability when facing non-critical image noise.

\noindent\textbf{3) Severely Suboptimal Benchmark ($T_{severe}$):}
Comprising 514 Level 3 images.
This set reflects  \textit{extreme acquisition constraints}, originating from difficult clinical situations where patient cooperation is limited, \textit{e.g.}, severe dyspnea or recumbent immobility.
This benchmark tests the model's ability to disentangle diagnostic signals from significant artifacts in complex cases.

\noindent\textbf{4) Auxiliary Benchmark ($T_{aux}$):}
Containing 669 naturally occurring low-quality samples from the official test split.
It serves as an external reference for uncurated, real-world quality shifts.

\subsection{Evaluation Metrics}

To ensure a holistic assessment, we evaluate model performance from both linguistic and clinical perspectives.
For Natural Language Generation (NLG), we employ standard metrics including \textbf{BLEU-1} \cite{papineni2002bleu}, \textbf{BLEU-4} \cite{papineni2002bleu}, \textbf{METEOR} \cite{banerjee2004meteor}, and \textbf{ROUGE-L} \cite{lin2005recall}.
These metrics quantify the lexical fluency and n-gram overlap between the generated hypotheses and ground-truth references.
While effective for measuring textual coherence, these standard metrics often fail to capture the precise medical correctness required for diagnostic reporting, necessitating a more specialized clinical evaluation.

Consequently, we prioritize Clinical Efficacy (CE) by leveraging the \textbf{RadEval} \cite{xu2025radeval} toolkit, a comprehensive framework designed for medical text evaluation.
We first utilize the internal CheXbert labeler to extract 14 standard pathology observations, calculating \textbf{Precision}, \textbf{Recall}, and \textbf{F1 scores} based on diagnostic label alignment.
To capture deeper semantic consistency beyond simple classification, we incorporate advanced metrics supported by RadEval: \textbf{RaTEScore} \cite{zhao2024ratescore} assesses entity-attribute alignment using medical encoders, \textbf{RadGraph} \cite{jain2021radgraph} evaluates the preservation of complex clinical relationships through knowledge graph construction.

\subsection{Baselines}

To strictly investigate the impact of image quality without architectural interference, we selected two representative foundation models for our fine-tuning experiments: \textbf{R2GenGPT} \cite{wang2023r2gengpt} and \textbf{Qwen3-VL} \cite{yang2025qwen3}.
R2GenGPT represents the specialized, lightweight biomedical paradigm, while Qwen3-VL represents the powerful, generalist Large Vision-Language Model (LVLM) paradigm.
We deliberately chose these streamlined architectures over complex, module-heavy designs (e.g., those with external memory banks or graph aligners).
This minimalist selection strategy allows us to isolate data quality as the primary variable, minimizing confounding factors arising from intricate architectural priors. R2GenGPT is trained using the officially recommended parameters. We fine-tuned the \textbf{Qwen3-VL-4/8B-Instruct} model. We employed the LoRA (Rank=64, Alpha=128), targeting all linear modules while fully training the visual resampler. We optimized the model using a learning rate of $1\times 10^{-4}$ with a cosine scheduler and a warmup ratio of 0.03. Training was conducted for 2 epochs with a per-device batch size of 4 and a gradient accumulation step of 1 under BF16 precision.

Furthermore, to validate the effectiveness of our base models, we extensively benchmarked them against established state-of-the-art methods on the $T_{std}$ and $T_{aux}$ sets.
Comparison targets include classical models like MambaXray and recent transformers like EKAGen.
Our fine-tuned baselines consistently achieve Tier-1 performance on these standard benchmarks.
This confirms that any performance degradation observed in the stress tests ($T_{mild}$ and $T_{severe}$) stems from the intrinsic difficulty of the quality shifts rather than insufficient model capacity.

\subsection{Results}
\noindent\textbf{Overall Performance Analysis.} As presented in Table~\ref{tab:main_results}, we conducted a comprehensive evaluation against both generalist foundation models and specialized state-of-the-art methods. To validate the efficacy of our backbone, we first established a performance baseline on the Standard benchmark $\mathcal{T}_{std}$. In this regime, the base R2GenGPT model delivers competitive results that are on par with specialized architectures such as Mamba\_Xray, confirming that our foundation is sufficiently strong for rigorous comparison. The significant advantage of our proposed framework emerges when handling real-world quality shifts. Unlike existing methods that suffer performance degradation on uncurated data, our Dual-loop Training Strategy demonstrates superior generalization. This is particularly evident on the Auxiliary benchmark $\mathcal{T}_{aux}$, where our approach achieves an F1 score of 0.307, effectively outperforming the task-specific EKAGen model which scores 0.299. These results indicate that while current state-of-the-art models are highly optimized for ideal inputs, our method successfully learns quality-invariant diagnostic features that are essential for robust clinical application.

\paragraph{Effect of the Proposed Dual-loop Training Strategy.}
We evaluate the impact of our Dual-loop Training Strategy (DTS) by comparing R2GenGPT-DTS with a baseline trained without it. The results show that DTS yields consistent gains on degraded inputs. Specifically, on $T_{mild}$, DTS significantly improves clinical precision from 0.353 to 0.461. It also boosts the F1 score from 0.291 to 0.321. Improvements in RaTEScore and RadGraph further demonstrate robustness to mild artifacts. On $T_{severe}$, DTS continues to enhance both NLG and CE metrics, such as BLEU-4 and RadGraph score. These results validate our motivation. They suggest that DTS encourages gradient coherence, helping the model focus on quality-invariant diagnostic evidence rather than quality-specific cues.

\section{Conclusion}

In this work, we challenged the prevailing assumption of ideal image quality in medical vision-language modeling. We proposed a robust chest X-ray report generation framework that explicitly accounts for the inevitable degradations found in clinical settings. Through the development of the AQAA and the LRRG benchmark, we exposed the fragility of current systems and provided a standardized method for robustness evaluation. Furthermore, our Dual-loop Training Strategy with gradient consistency successfully enables quality-agnostic feature learning. Our results empirically validate that robustness does not have to come at the cost of standard performance. By bridging the gap between clean research datasets and noisy real-world scans, our approach paves the way for more trustworthy and safe clinical assistance agents.


\section{Limitations}
Despite the promising results achieved by our framework, we acknowledge several limitations that require further investigation: First, the proposed Dual-loop Training Strategy introduces additional computational overhead during training due to the bi-level optimization, although inference latency remains unaffected. Second, the performance of our AQAA relies on the pre-trained VLM foundation, which may inherit potential biases or overlook rare, domain-specific artifacts. Third, our evaluation is currently restricted to 2D chest X-rays within the MIMIC-CXR dataset; further validation is required to confirm generalizability to 3D modalities (e.g., CT, MRI) and multi-center settings.

\section{Ethical Considerations}
We strictly adhered to the data use agreements of MIMIC-CXR, ensuring all data remained de-identified. We emphasize that our system is an assistive tool; AI-generated reports should always be verified by clinicians to prevent potential harm from hallucinations. Furthermore, our work contributes to healthcare equity by addressing performance disparities on low-quality images, which are often prevalent in critical care and resource-constrained environments.



\bibliography{ref}

\appendix




\end{document}